%% file: DMSLSTM-article-philipp.tex
\documentclass[a4paper,fleqn]{cas-dc}

\usepackage[numbers]{natbib}
\usepackage{subcaption}
\usepackage{xcolor}
\usepackage{soul}
\usepackage{graphicx}
\usepackage{makecell} 

\newcommand{\red}[1]{\textcolor{red}{#1}}

\definecolor{delete}{gray}{0.8}

\def\tsc#1{\csdef{#1}{\textsc{\lowercase{#1}}\xspace}}
\tsc{WGM}
\tsc{QE}

\begin{document}
\let\WriteBookmarks\relax
\def\floatpagepagefraction{1}
\def\textpagefraction{.001}

\shorttitle{Classification of Differential Mobility Spectrometer's Dispersion Plots with LSTM Neural Networks}    

\shortauthors{Rauhameri et al}

\title [mode = title]{Classification of Volatile Organic Compounds by Differential Mobility Spectrometry Based on Continuity of Alpha Curves}  




\author[1]{Anton Rauhameri\corref{1}}[orcid=0000-0002-7021-7868]
\cormark[1]
\cortext[1]{Corresponding author}
\credit{Conceptualization, Methodology, Software, Validation, Formal analysis, Investigation, Data Curation, Writing - Original Draft, Visualization}
\ead{anton.rauhameri@tuni.fi}

\author[2]{Angelo Robiños}[orcid=0009-0003-9200-9909]
\credit{Investigation, Writing - Review \& Editing}
\author[3]{Osmo Anttalainen}[orcid=0000-0001-8518-0407]
\credit{Writing - Original Draft, Writing - Review \& Editing }
\author[1]{Timo Salpavaara}[orcid=0000-0002-6331-6972]
\credit{Writing - Original Draft, Resources}
\author[1]{Jussi Rantala}[orcid=0000-0002-6308-7874]
\credit{Methodology, Writing - Review \& Editing}
\author[1]{Veikko Surakka}[orcid=0000-0003-3986-0713]
\credit{Conceptualization, Resources, Writing - Review \& Editing, Supervision, Project Administration, Funding Acquisition}
\author[1]{Pasi Kallio}[orcid=0000-0002-2890-6782]
\credit{Resources, Writing - Review \& Editing, Funding Acquisition}
\author[1]{Antti Vehkaoja}[orcid=0000-0003-3721-3467]
\credit{Conceptualization, Resources, Writing - Original Draft, Writing - Review \& Editing, Supervision, Project administration, Funding acquisition}
\author[1]{Philipp Müller}[orcid=0000-0003-4314-7339]
\credit{Conceptualization, Validation, Writing - Original Draft, Writing - Review \& Editing, Supervision, Project administration}





\affiliation[1]{organization={Tampere University},
	addressline={Korkeakoulunkatu 7},
	city={Tampere},
	postcode={33720},
	country={Finland}}

\affiliation[2]{organization={Åbo Akademi University},
	addressline={Henriksgatan 2},
	city={Turku},
	postcode={20500},
	country={Finland},
}

\affiliation[3]{organization={Olfactomics Oy},
	addressline={Korkeakoulunkatu 7},
	city={Tampere},
	postcode={33720},
	country={Finland}
}

%









\begin{abstract}		
Background: Classification of volatile organic compounds (VOCs) is of interest in many fields. Examples include but are not limited to medicine, detection of explosives, and food quality control. Measurements collected with electronic noses can be used for classification and analysis of VOCs. One type of electronic noses that has seen considerable development in recent years is Differential Mobility Spectrometry (DMS). DMS yields measurements that are visualized as dispersion plots that contain traces, also known as alpha curves. Current methods used for analyzing DMS dispersion plots do not usually utilize the information stored in the continuity of these traces, which suggests that alternative approaches should be investigated.
	
	Results: In this work, for the first time, dispersion plots were interpreted as a series of measurements evolving sequentially. Thus, it was hypothesized that time-series classification algorithms can be effective for classification and analysis of dispersion plots. An extensive dataset of 900 dispersion plots for five chemicals measured at five flow rates and two concentrations was collected. The data was used to analyze the classification performance of six algorithms. According to our hypothesis, the highest classification accuracy of 88\% was achieved by a Long-Short Term Memory neural network, which supports our hypothesis.
	
	Significance: A new concept for approaching classification tasks of dispersion plots is presented and compared with other well-known classification algorithms. This creates a new angle of view for analysis and classification of the dispersion plots. In addition, a new dataset of dispersion plots is openly shared to public.
\end{abstract}
%


\begin{highlights}
\item A novel method for classification dispersion plots is presented 
\item Application of time series models on dispersion plot classification is proposed
\item A data set containing dispersion plots of 5 different chemicals measured at 5 different flow rates is presented
\end{highlights}


\maketitle

\section{Introduction}\label{sec:introduction}
\input{tex/introduction-philipp.tex}








\section{Data acquisition}\label{sec:acquisition}
\input{tex/datagather-philipp.tex}

\section{Data description}\label{sec:ddescription}
\input{tex/ddescription-philipp.tex}

\section{Classification methods}\label{sec:cmethods}
\input{tex/classification_algos-philipp.tex}

\section{Results and Discussion}\label{sec:results}

\input{tex/results-philipp.tex}

\section{Conclusions and outlook}\label{sec:conclusions}
\input{tex/conclusions-philipp.tex}

\section*{Appendix}
\red{Supplementary material related to this article can be found online at LINK}

\section*{Acknowledgement}
This work was supported by Academy of Finland under grants 323498, 323529 and 323530 and Suomen Kulttuurirahasto under grant 50221583. Authors thank Tuan-Anh Tran for his help in collecting the data.
Author Osmo Anttalainen is a shareholder of Olfactomics Ltd, a company that developed IonVision - differential mobility spectrometer. The other authors declare no competing interests.
\printcredits

\bibliographystyle{model1-num-names}

\bibliography{zotero}



\end{document}

%% file: tex/introduction-philipp.tex
Classification of scents is of interest in many fields. Notable application areas include but are not limited to medicine, food quality control, detection of warfare agents, and digitalization of scents (see e.g.\cite{wilson2009} and references therein). For example, in the medical field, scents emitted by evaporated tissue can be used for distinguishing pathological and healthy tissue \cite{Lubes2018}, \cite{Kontunen2021} and identifying the tumor type during surgical operation \cite{haapalamethod2022}. Likewise, in food quality control scents can be used for checking the maturity level of fruits and vegetables or ensuring freshness of meat products \cite{hernandez-mesa_current_2017}.
Accurate classification and analysis of scents will open new possibilities for, for example, diagnosing cancer patients \cite{Haapala2019},
and digitalizing and transmitting scents over the internet \cite{panagiotakopoulosdigital2022}.

An emerging field for applying classification of scents relates to extended reality (XR) research and development. XR applications aim at creation of digital twins of the real world with which humans can interact as if in a real world \cite{archer_odour_2022}. For such developments measurement and analysis of the outside world is important, for example, for remote work and operations. Measurement and classification of real-world scents is needed in order to be able to reproduce them with the help of olfactory displays (OD) in virtual reality (e.g., \cite{nieminen_compact_2018}, \cite{salminen_olfactory_2018}, \cite{muller_online_2019}). The present work is part of a more extensive project focusing on sensing and reproducing scents to virtual reality. We are developing scent classifications, olfactory displays, and VR environments. The present work focuses on improving classification of five scents to enable us using the full potential of the five channels in our OD prototype.

Various approaches, for sensing scents, exist that often rely on detecting and analyzing volatile organic compounds (VOC). VOCs in air can be separated and measured due to their different physical properties. One of the separation methods for VOCs is Differential Mobility Spectrometry (DMS). Several manufacturers produce commercial DMS devices. In this study the IonVision from Olfactomics \cite{olfactomics} was used. The sample in the IonVision is ionized with 4.9 keV photons from x-ray source in a sequence of atmospheric pressure reactions. The sequence is understood to start from electron extraction from $N_2$ and $O_2$ molecules and interaction between neutral molecules in air leading quickly to the formation of protonated water clusters $(H_2O)_nH^+$ called positive reactant ions (RIP), and  $O_2^-(H_2O)_n$ called negative reactant ions (RIN) \cite{good_mechanism_1970}. When the reactant ions collide with neutral analyte molecules, product ions are generated. The formation of product ions depends on the concentration of reactant ions and neutral analyte, as well as the proton affinity of the reactants. The higher the proton affinity, the more probable is the formation of such product ions.
The ionized molecules enter then a drift region, which acts as a filter. In this region, ions are carried by neutral gas flow and are separated in asymmetrically oscillating electric field based on their field dependence. The passband of the DMS ion filter is controlled by a separation field which is defined by an amplitude of oscillating voltage, and compensation field, which is defined by DC component added to  the oscillating field. The ions that pass the filter are detected in Faraday plate detector following immediately after the drift region. The DMS filter operates by scanning both separation and compensation fields. DMS is often used as a sample pre-separation method in mass spectrometer analysis or as a detector in Gas Chromatography Spectroscopy (see e.g. \cite{martinelli_proposal_2016} and \cite{coy_detection_2010}).

However, DMS can be also used independently for VOC analysis \cite{lepomaki_laser_2022}. A measurement of DMS is often represented as a matrix of numbers, where each row represents measurements with a fixed separation voltage $U_{sv}$ and varying compensation voltages $U_{cv}$. When visualized, the matrix is commonly referred to as a dispersion plot. It is important to note that Ion mobility spectrometry, including DMS, is an experimental technique and without prior testing it is impossible to infer from the dispersion plot what chemical was measured. Instead, prior knowledge of the sample as well as a pre-trained classification algorithm is necessary to identify the measured chemical with the dispersion plot as input. Various machine learning methods have been used recently for the classification of chemicals based on dispersion plots. Lepomäki et al. \cite{lepomaki_laser_2022} used shrinkage linear discriminant analysis (sLDA), support vector machines (SVM), and a convolutional neural network (CNN) to differentiate between five porcine tissue types introduced to DMS through laser desorption. The classification accuracies were 79.8\% (sLDA), 79.0\% (SVM), and 86.4\% (CNN). The authors also demonstrated that, intensity-wise, the mean and standard deviation of the dispersion plots for each type of tissue had noticeable visual differences in intensity. Hence, the authors already expected high classification accuracy for CNN. However, in general, dispersion plots of different VOCs cannot be distinguished by visual inspection alone, making the accurate classification more challenging than what was demonstrated in \cite{lepomaki_laser_2022}.

In a recent study conducted by Haapala et. al. \cite{haapalamethod2022}, Linear Discriminant Analysis (LDA) achieved an average accuracy of 85\% in binary classification of isocitrate dehydrogenase (IDH) Mutation in Gliomas based on dispersion plots. However, their dataset consisted of only 352 samples, which made it challenging to apply more complex classification algorithms. With such limited data, it is difficult to develop a neural network that provides accurate classification.

There are many other works trying to classify or analyze DMS dispersion plots with the help of machine learning algorithms . For example, \cite{fowler_neural_2023} explored a comparative analysis of neural network platforms by classifying VOCs with tandem DMS. In their work, the protonated monomers were isolated by applying a strong electric field and analyzed at the second DMS filter aiming to create fragments, and using the fragments as basis for the neural network analysis. Although, the setup differs significantly from the one presented in this paper, the classification accuracy obtained with the neural network developed by the authors was over 90\% for familiar compounds and 64\% for unfamiliar. As opposed to work presented in \cite{fowler_neural_2023}, only one DMS filter was used in the current work. Our setup is simpler and produce mostly native product ions and not field induced fragments. Additionally, using the tandem DMS for that specific application suggests that the authors had some prior information about the samples being analyzed.

Another work by \cite{ieritano_predicting_2021} aimed at predicting dispersion curves yielded by DMS. The authors used a Random Forest regression algorithm, and achieved high accuracies\footnote{Authors measured accuracy by the Mean Absolute Error (MAE).} for different types of alpha curves. Their setup and sample preparation methods, however, differed significantly from the work presented in this paper. In \cite{ieritano_predicting_2021} DMS was only used as a filter before measurements were taken by a mass spectrometer. Furthermore, nitrogen was used as carrier gas. This has the advantage, compared to using ambient air as carrier gas, that nitrogen contains less trace chemicals reducing contamination. The disadvantage is, however, that it requires higher electric fields. Finally, the authors knew in advance mass-to-charge ratio (mz) and collision cross-section (CCS), which simplifies the prediction task considerably.

Rajapakse et\,al. \cite{rajapakse_automated_2018} demonstrated a promising method for classification of dispersion plots with partial Least squares - discriminant analysis. They showed that the first two principal components represented over 65\% of variability in the dispersion plots and are able reveal clear clusters of chemicals. However, compared to the work presented in this paper, the samples in \cite{rajapakse_automated_2018} were prepared using pure nitrogen as dilution gas.

The works mentioned above, and many other, demonstrate development of DMS methods and high interest of algorithmic analysis of the measurement results. Different research groups use various DMS techniques for obtaining reliable results of VOC analysis. However, all these methods differ significantly in their setups, sample preparations, etc. This work is aimed to present a classification algorithm applied to a simple setup with weakly controlled environmental conditions. Additionally, the dataset is published to enable interested readers to compare their methods with the methods presented in this work.

DMS data, in general, have high dimensionality. Measurements for tens or even hundreds of different separation and compensation voltages can be collected for only a single dispersion plot. Therefore, using regularization \cite{friedman_regularized_1989} and principal component analysis \cite{altman_curses_2018} to avoid the curse of dimensionality is suggested. In this paper, we explore an alternative approach to address the curse of dimensionality. Our hypothesis is that  dispersion plots can be interpreted as a set of sequential measurements. In order to verify our hypothesis, a new dataset containing five chemical solutions measured at five carefully controlled flowrates was collected. This was done in order to produce variation and generate multilabel data. Based on the hypothesis, for example, a 40-by-200 matrix can be interpreted as sequence with 40 measurements \footnote{Which means each row is one measurement of 200 features and each column a sequence of 40 measurements for one feature}, where each measurement is of dimension 200. This means that at each step, values for 200 features are available. As a result, there is no need to apply PCA and lose the information preserved between measurement points. However, ion separation is independent of the order of separation and compensation field sequences, and the sequence could be run in random order or using only a limited amount of $U_{sv}$-$U_{cv}$ pairs to save data acquisition time. For convenience of configuration, parameters are usually scanned in increasing order.

The contribution of this article is threefold. First, new results are presented for classifiers that at their best can distinguish between the different chemical dilutions measured at different flow rates with an accuracy of 89\% on the collected data set. To our knowledge, there are no carefully controlled studies that reported accuracies of 89\% or higher for multilabel machine learning problems using DMS measurements. Second, it is demonstrated how time-series deep learning networks (Long-Short Term Memory, LSTM) can be applied to DMS data, and that they outperformed other well-known classifiers. Third, the collected data set is shared openly to enable anyone interested to repeat our tests and compare their own classifiers with the algorithms proposed in this paper.
In section \ref{sec:acquisition} the data gathering process is described. Section \ref{sec:ddescription} describes the collected data set. Section \ref{sec:cmethods} briefly describes algorithms used for classification of DMS dispersion plots. Section \ref{sec:results}  discusses results of the applied methods. In Section \ref{sec:conclusions}, an outlook and plans for further improvements are discussed.

%% file: tex/datagather-philipp.tex
As discussed earlier, our current olfactory display prototype can currently emit mixtures of up to five chemicals, thus, five chemicals were selected that each on their own have a distinct scent that could be used in a VR without mixing it with other chemicals.
For example, ethyl 2-methylbutyrate and (-)-Carvone smell like strawberry and mint respectively. Thus, the analysis in this paper was limited to five chemicals.


\subsection{Materials}
R-(\textendash)-carvone ($C_{10}H_{14}O$, 98\%), ethyl-2-methylbutyrate ($C_{7}H_{14}O_2$, 99\%), methyl cyclopentenolone ($C_{6}H_{8}O_2$, $\ge$ 98.0\%), 2-phenylethanol ($C_{8}H_{10}O$, $\ge$ 99.0\%), n-butanol ($C_{4}H_{10}O$, 99.9\%), and propylene glycol ($C_{3}H_{8}O_2$, $\ge$ 99.5\%)  from Merck were used as stimuli for DMS measurements. Except for methyl cyclopentenolone, which is solid, all chemicals were in liquid form. n-Butanol served as the reference substance while propylene glycol was used to dilute the concentrated scents. All chemicals used in this work are presented in Table \ref{tab:chemicals}.

\subsection{Sample preparation and measurement}

The experiment utilized containers and apparatus that were thoroughly cleaned to prevent contamination of the samples. Solutions of percent volume (\%v/v) and percent weight (\%w/w) concentrations were prepared from liquid and solid substances, respectively. In the case of undissolved solid chemicals, the flasks were sonicated until virtually no solids remained. Two concentration levels per chemical were then prepared as samples: 1/100 (1\% v/v or w/w) for a stronger and 1/10\,000 (0.01\,\% v/v or w/w) for a milder scent.

\begin{figure}[h!]
	\centering
	\includegraphics[width=0.3\textwidth]{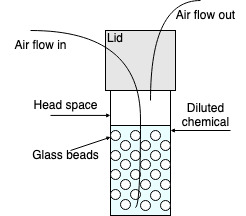}
	\caption{Schematic representation of chemical put into the vial}
	\label{fig:vial_sketch}
\end{figure}

A custom-built odor display \cite{sigchi} was used to vaporize the sample solutions, to adjust the intensity of the odor gas by mixing it with clean air and to deliver the diluted odor flow to the DMS. In short, filtered air was pumped into two channels (odor and dilution air) and the air flows were adjusted using two mass flow controllers (MFCs). A scented airflow was created by passing a known volume flow of air through an odor source. This channel could be combined with the dilution air flow (five standard liters per minute (SLPM)) with a three-way valve. The DMS was connected in parallel to the output of the system to take in samples of scented airflow.
A class vial filled with 2 mm diameter glass beads and odorant solution as in Figure \ref{fig:vial_sketch}, acted as odor source.
The beads were added to make the air bubbles travelling from the bottom of the vial spend an equal amount of time soaking up the analyte from the solution before measurement. Schematically this strategy aimed to improve the repeatability of the results.

The samples were measured in random sequence, with increasing flowrates, to account for instrument drift. During one day five sets of chemicals were measured. Each set contained one chemical measured at 8, 16, 32, 64 and 128 sccm flow rates mixed with room air. Thus, each chemical at a certain flow rate was measured once per day. The procedure was repeated on 25 days. Propylene glycol was also measured in every five sets of samples to monitor the baseline and to check for any cross-contamination. Each set of chemicals was repeatedly measured for five days to further account for within-day fluctuations in the odor display, temperature and humidity, which could affect concentration.


\begin{table}
\caption{List of used chemicals, their abbreviations and Chemical Abstract Service (CAS) Registry Numbers}
\label{tab:chemicals}
\begin{tabular}{c|c|c|c}
	\textbf{Chemical} & \textbf{Abbreviation} & \textbf{CAS} & \textbf{Purity \%}\\ 
	\hline
	$C_{10}H_{14}O$ & Carvone & 6485-40-1 & $\ge$ 98.0\\ 
	$C_{7}H_{14}O_2$ & E2MB & 7452-79-1 & $\ge$ 99.0\\ 
	$C_{6}H_{8}O_2$ & MCP & 80-71-7 & $\ge$ 98.0\\ 
	$C_{8}H_{10}O$ & 2PEtOH & 60-12-8 & $\ge$ 99.0\\ 
	$C_{4}H_{10}O$ & nBuOH & 71-36-3 & $\ge$ 99.9
\end{tabular} 
\end{table}

%% file: tex/ddescription-philipp.tex

\begin{figure}[!h]
	\begin{subfigure}{0.5\textwidth}
		\includegraphics[width=1\textwidth]{"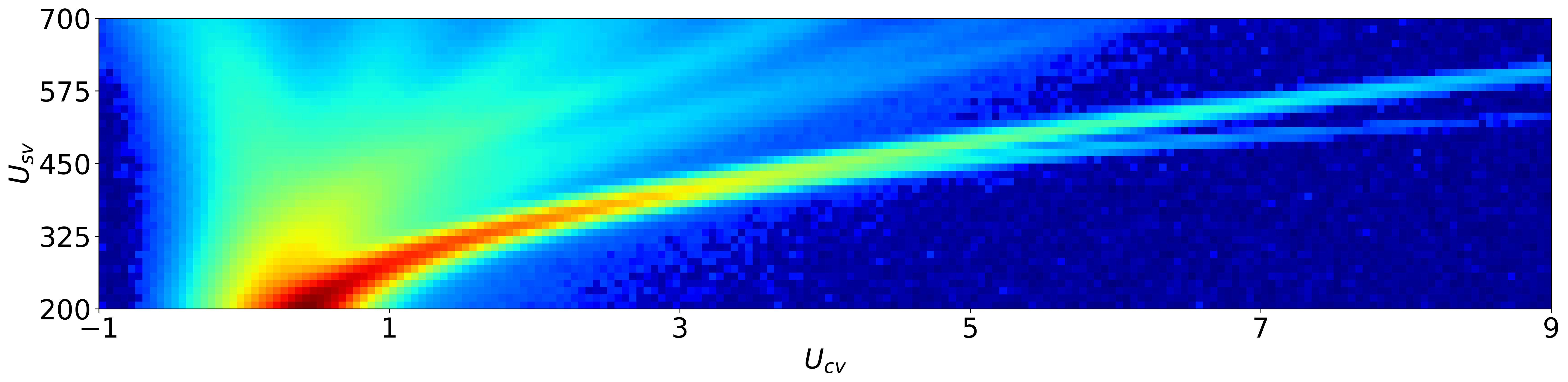"}
		\caption{Dispersion plot of n-Butanol}
		\label{fig:nbuohexample}
	\end{subfigure}
	\begin{subfigure}{0.5\textwidth}
		\includegraphics[width=1\textwidth]{"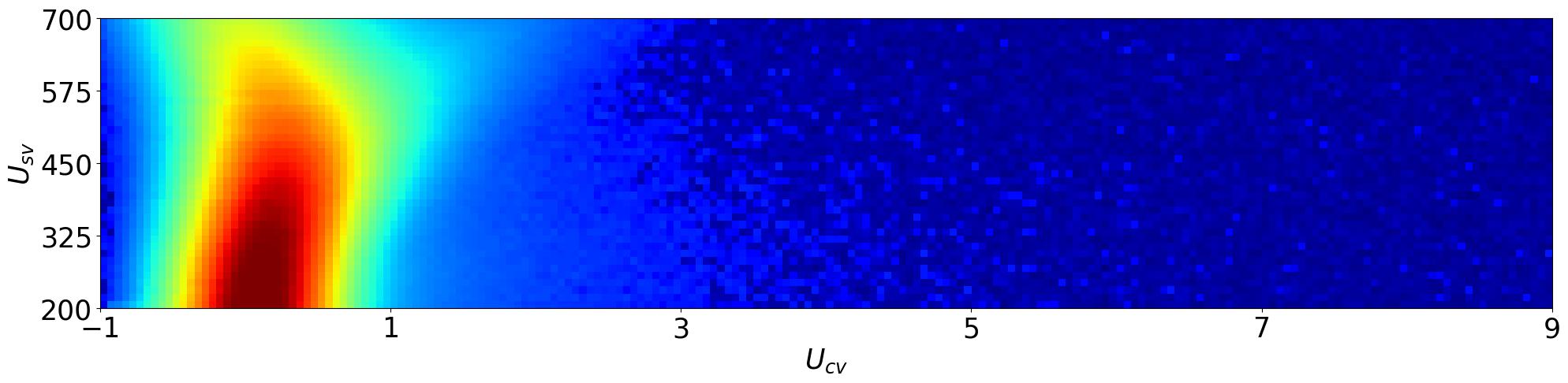"}
		\caption{Dispersion plot of ethyl-2-methylbutyrate}
		\label{fig:e2mbexample}
	\end{subfigure}
	\caption{Examples of dispersion plots of two chemicals, n-Butanol and ethyl-2-methylbutyrate showing positive ions, measured by Olfactomics IonVision device. For better visualization cubic root was applied to the raw values. The typical pattern in dispersion plots of all chemicals, except those of ethyl-2-methylbutyrate, is shown in (a). Ethyl-2-methylbutyrate patterns in dispersion plots resembled subfigure (b).}
	\label{fig:dplotsexample}
\end{figure}

A dispersion plot (example is shown in Figure \ref{fig:dplotsexample}) represents a matrix of size $U_{sv} \times U_{cv}$, where rows represent compensation voltages for a fixed separation voltage $U_{sv}$ and columns represent separation voltages for a fixed compensation voltage $U_{cv}$. Since the optimal measurement parameters were not known in advance, the default values proposed by the IonVision developer were used, meaning that separation voltage ranged from 200\,V to 700\,V, and compensation voltage ranged from -1\,V to 9\,V.
The frequency of the separation voltage was 1\,MHz with 20\% duty cycle and rectangular waveform. The gap width of the DMS sensor\footnote{Increasing the gap decreases the electric field. Here a gap of 0.25\,mm was used to obtain a high electrical field with practical low electrical potential} used was d=0.25 mm, leading to a high electric field strength of $(D-1)U_{sv}\,d^{-1}$ and a low field strength of $-D\,U_{sv}\,d^{-1}$, where $D$ is the duty cycle\footnote{The low and high field strengths are thus 1600~V/cm and 22400 V/cm respectively.}. Duty cycle in IonVision is tunable, but here it was 22\%.

\begin{figure}[h]
	\begin{subfigure}{0.5\textwidth}
		\includegraphics[width=1\textwidth]{"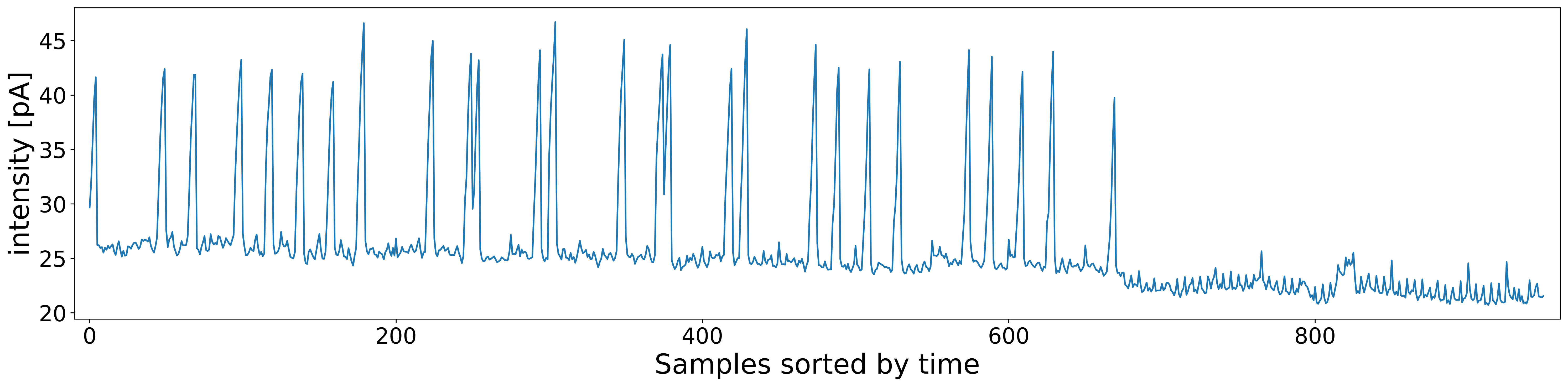"}
		\caption{including ethyl-2-methylbutyrate}
		\label{fig:drift_inc_e2mb}
	\end{subfigure}
\begin{subfigure}{0.5\textwidth}
	\includegraphics[width=1\textwidth]{"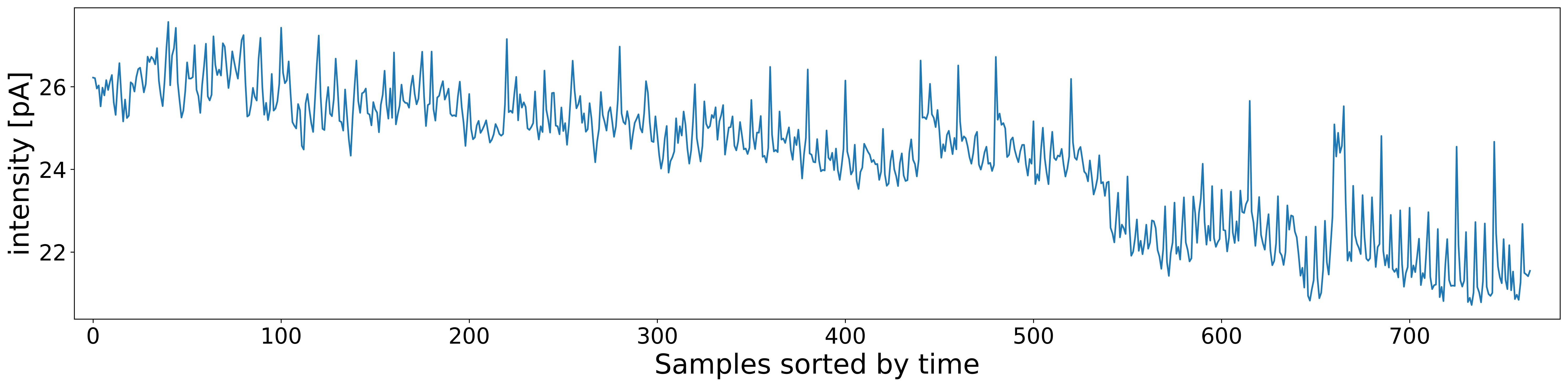"}
	\caption{without ethyl-2-methylbutyrate}
	\label{fig:drift_exc_e2mb}
\end{subfigure}
\caption{Drift of the average intensity over all compensation voltages at $U_{sv}$ = 200\,V.}
\label{fig:drifts}
\end{figure}

In all measurements, of all chemicals, drift in the DMS device output ion current values (referred later as intensity) was observed.
 Figure \ref{fig:drifts} shows the drift of the measurements ordered by time. The spikes in Figure~\ref{fig:drift_inc_e2mb} are measurements of ethyl-2-methylbutirate.
%

\begin{figure}[!h]
	\centering
	\includegraphics[width=0.95\linewidth]{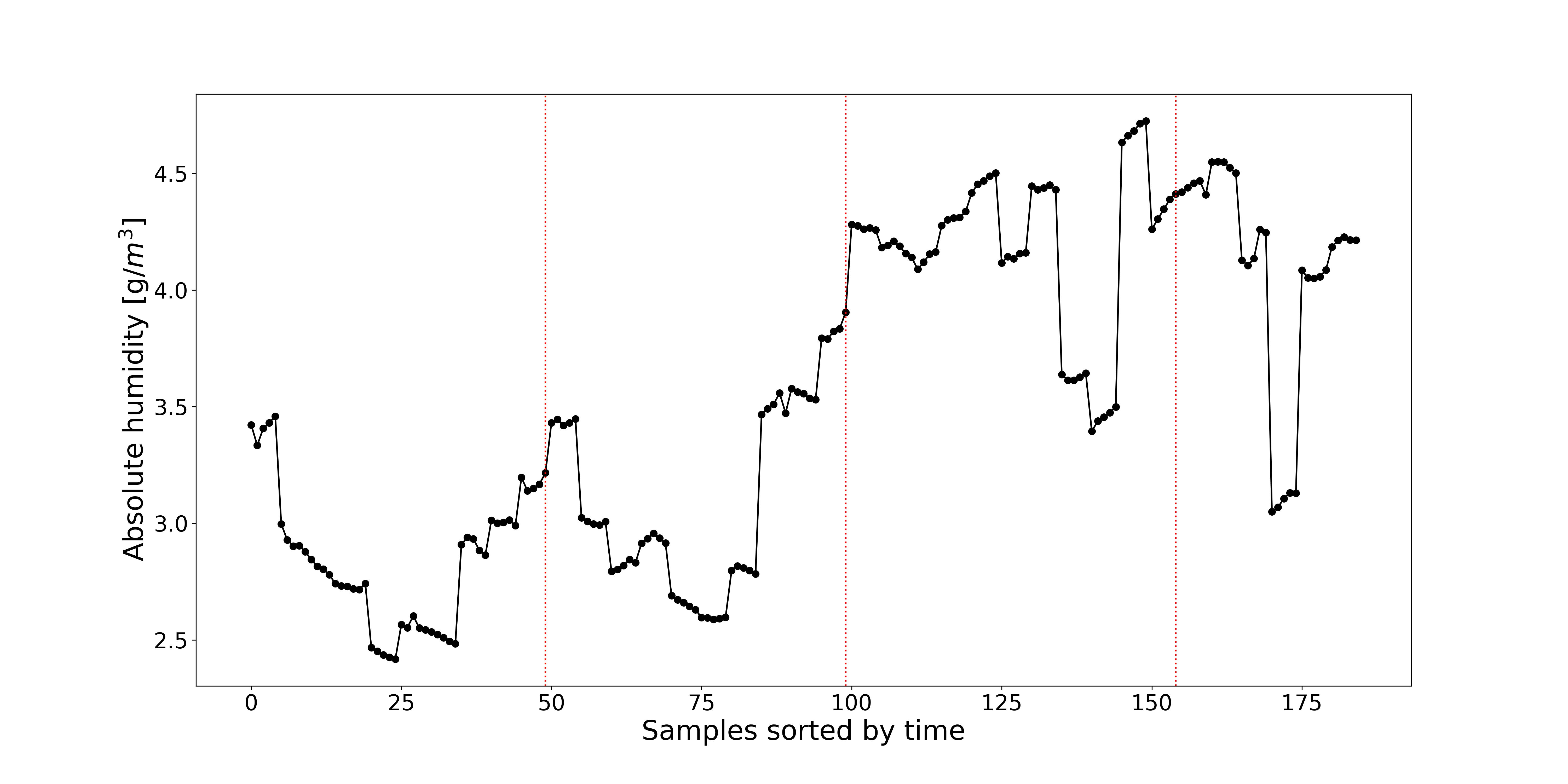}
	\caption{Moisture level of nBuOH during measurements. The red dotted lines depict moisture filter maintenance.}
	\label{fig:moisturedrift}
\end{figure}

\begin{figure}[h]
	\begin{subfigure}{0.5\textwidth}
		\includegraphics[width=1\textwidth]{"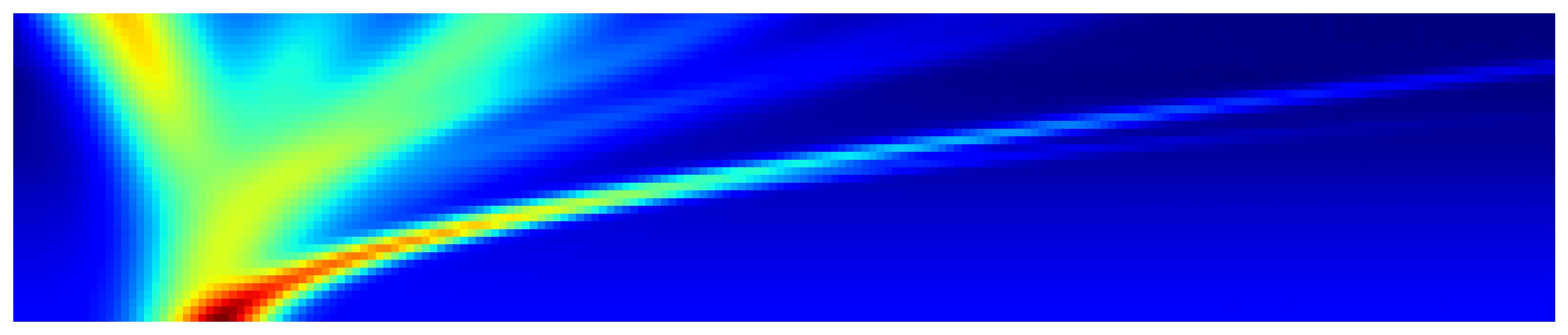"}
		\caption{Average dispersion plot of nBuOH}
		\label{fig:mean_nbuoh}
	\end{subfigure}
	\begin{subfigure}{0.5\textwidth}
		\includegraphics[width=1\textwidth]{"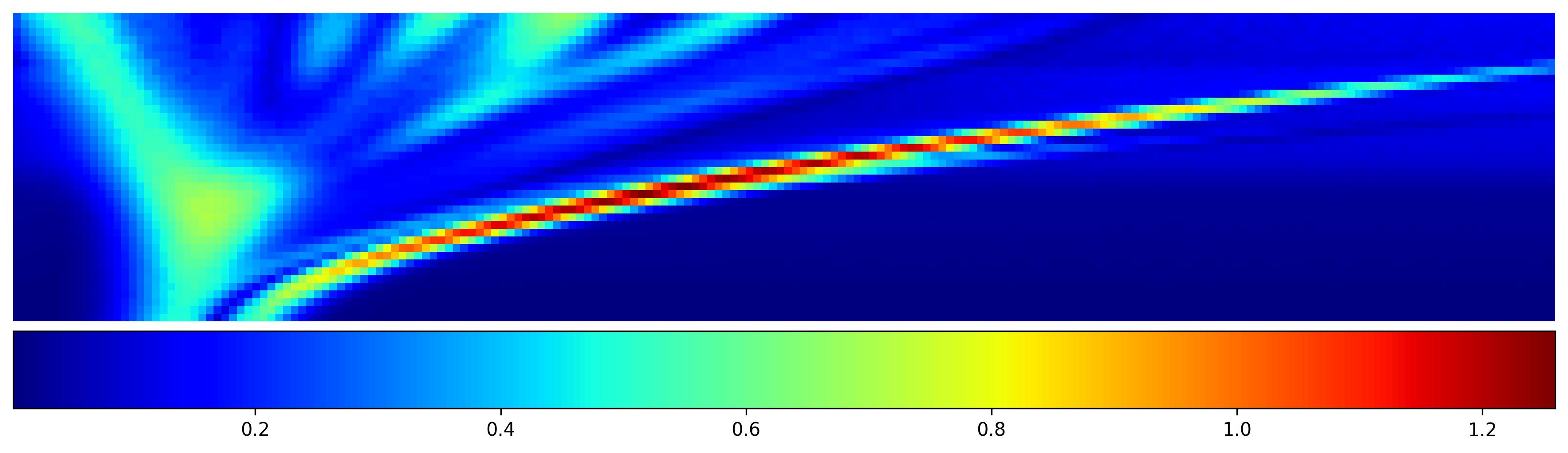"}
		\caption{Standard deviation dispersion plot of nBuOH}
		\label{fig:std_nbuoh}
	\end{subfigure}
	\caption{Dispersion plots showing (a) average appearance and (b) dissimilarities of nBuOH with the flowrate 32.}
	\label{fig:dissimilarities}
\end{figure}

Generally, all dispersion plots by visual inspection looked similar except for ethyl-2-methylbutyrate (E2MB). Dispersion plots of E2MB varied considerably from each other and Figure~\ref{fig:e2mbexample} shows one example. The dispersion plots of one chemical with the same flow rate cannot be identical on different days due to variable environmental factors. Figure \ref{fig:moisturedrift} shows humidity level for nBuOH during the measurement campaign. As can be seen, the humidity level trended during the measurement campaign. The humidity plays important role in DMS \cite{safaeidoctoral}. Additionally, standard deviation applied to a set of dispersion plots of nBuOH measured with the same flow rate reveals where they differed mostly (Figure \ref{fig:dissimilarities}). For generating Figure \ref{fig:dissimilarities} the dispersion plots of nBuOH of the same flow rate were stacked and mean and standard deviation were calculated pixelwise. The dispersion plots were normalized for better visualization. For visualization of other chemicals readers are referred to the supplementary material.

%% file: tex/classification_algos-philipp.tex
\subsection{Preprocessing}\label{sec:prepro}

All algorithms described in this section were implemented in Python using libraries scikit learn and Keras. In order to provide stable accuracy estimation, Repeated Stratified $k$-Fold Cross-Validation (RSCV) was used. The "stratified" in this abbreviation means maintaining a balanced ratio of classes in training and test sets. The "repeated" means that each fold is repeated $N$ times, and the average result is returned. The parameter $k$ defines the number of folds in the dataset and is different from the K used in the KNN. The $k$-value for RSCV was set to 5.

All dispersion plots in the collected dataset were scanned with a wide range of $U_{cv}$ (-1 to 9 V) and $U_{sv}$ (200 - 700 V) because no prior information was available on the ranges in which the dispersion plot would show reactions for the various chemicals. Consequently, the dispersion plots contain much redundant information. Anttalainen et al.~\cite{AnttalainenOsmo2020Pstu} described the different parts of a dispersion plot and proposed a method for determining the part of the dispersion plot in which separation of ions can be observed. The authors also stated that the Shannon entropy measure can be used to find the optimal section of a dispersion plot with a good signal-to-noise ratio.  

\begin{figure}[h]
	\begin{subfigure}{0.5\textwidth}
		\includegraphics[width=1\textwidth]{"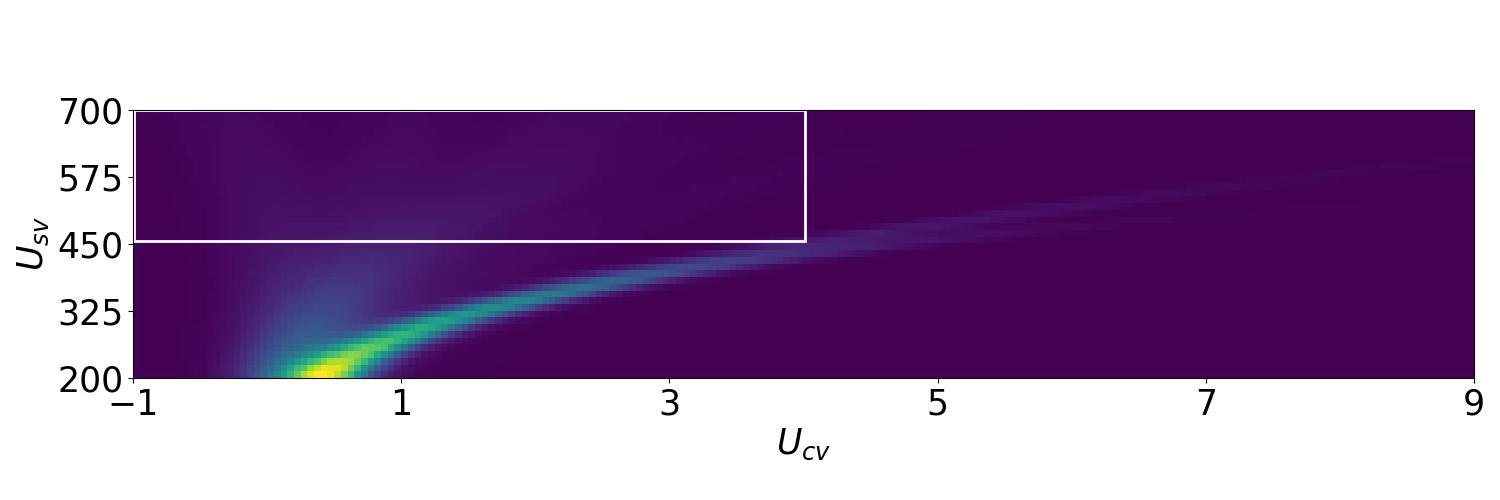"}
		\caption{}
		\label{fig:clippedplot}
	\end{subfigure}
	\begin{subfigure}{0.5\textwidth}
		\includegraphics[width=1\textwidth]{"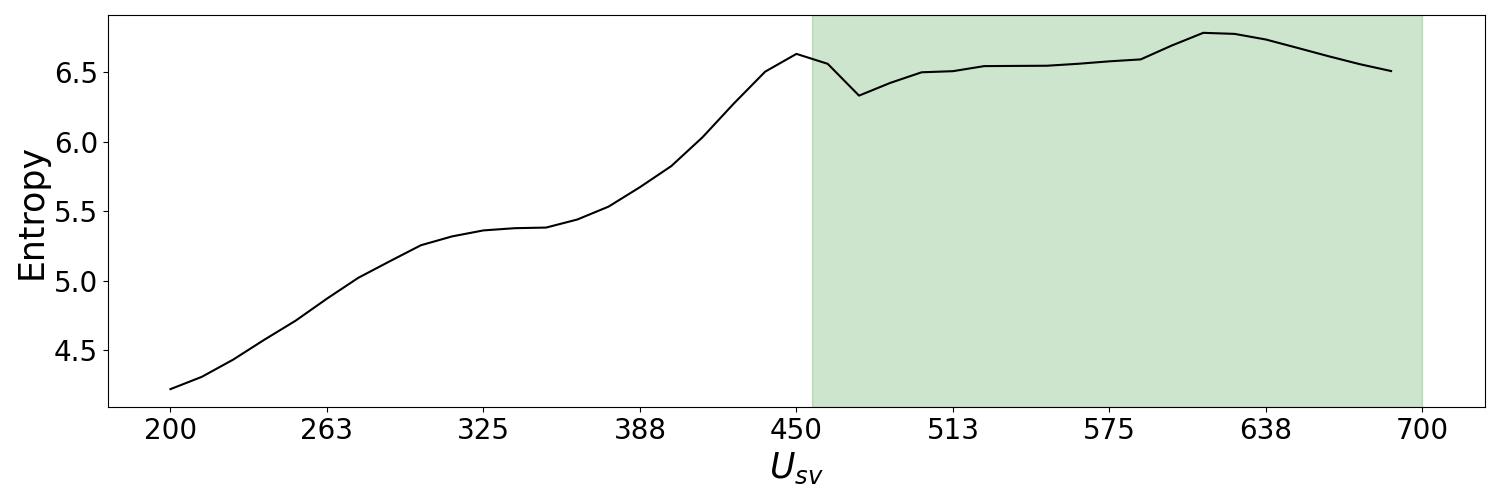"}
		\caption{}
		\label{fig:rowentropy}
	\end{subfigure}
	\caption{Reducing area of the dispersion plot used for analysis. Full measured dispersion plot of 2-phenylethanol is shown in (a). Row-wise entropy values of that dispersion plot are given in (b). The area filled with green shows the highest entropy. Based on (b) the part of the dispersion plot marked by the white rectangle in the upper left corner of (a) is selected for further processing.}
	\label{fig:entropias}
\end{figure}

From Figure \ref{fig:nbuohexample} it can be seen that the branching starts at high $U_{sv}$ and low $U_{cv}$ indicating that separation of ions has started. The analysis with Shannon entropy on a small selection of dispersion plots confirmed it and therefore dispersion plots were clipped
in a first preprocessing step to ranges of $U_{sv} =$ [456.41\,V, 700.00\,V] and $U_{cv}=$ [-1.00\,V, 3.97\,V]. Figure \ref{fig:clippedplot} shows the entropy metric applied to the dispersion plot of 2-phenylethanol (2PEtOH). Figure \ref{fig:rowentropy} shows row-wise entropy values. The green area on the plot shows that the top rows of a dispersion plot contain the most entropy. Thus, the top-left part of the dispersion plots was selected for further analysis.

The second preprocessing step was normalization of the data to ensure that all values were between -1 and 1. Since each dispersion plot represents a series of measurements with different separation voltages, the dispersion plots were normalized row-wise by subtracting the mean and dividing by the standard deviation. Originally, this method was proposed in~\cite{Virtanen_2022} but there rows were scaled to values between 0 and 1.

\begin{figure}[h]
	\centering
	\includegraphics[width=1\linewidth]{"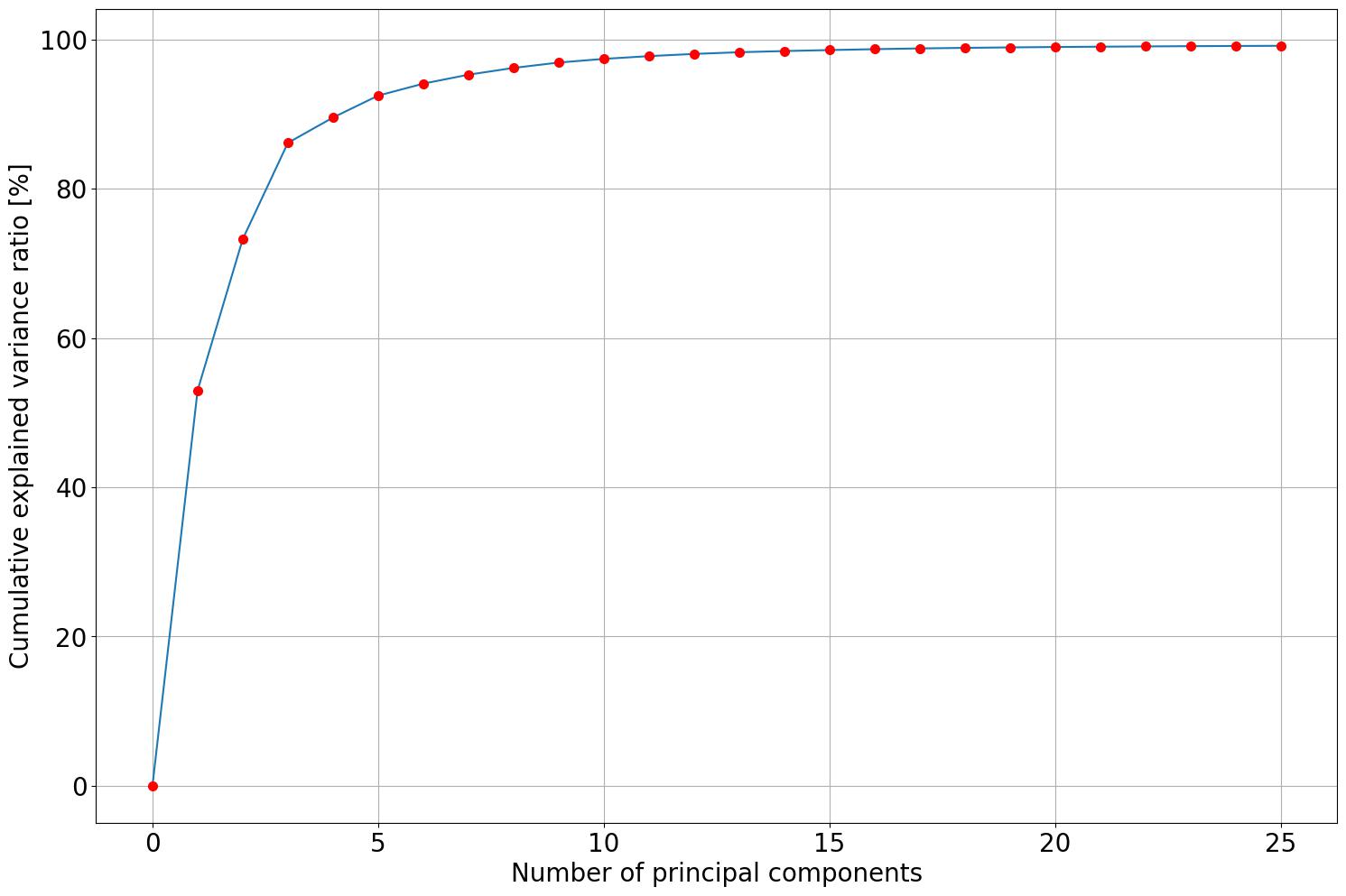"}
	\caption{Ratio of cumulative variance in original data explained by different numbers of principal components.}
	\label{fig:explainedvarianceratio}
\end{figure}

The data was compressed with principal component analysis (PCA) before applying KNN, LDA, ExtraTrees classifier, and MLP. PCA is a feature transformation technique that can be defined as the orthogonal projection of the data onto a lower dimensional linear space, known as the principal subspace, such that the variance of the projected data is maximized~\cite[p. 561]{bishop_pattern_2006}. In our tests applying PCA to the full data set resulted in the first 25 orthogonal principal components, out of the 2000 features in the data\footnote{Originally, a dispersion plot has the dimensionality of 8000. After clipping the dispersion plot with the method described previously, its dimensionality reduces to 2000. Finally, applyin PCA to 2000 features reduces the data further to 25 features.}, explaining over 99\% of the variability in the data (see Figure \ref{fig:explainedvarianceratio}). The PCA was used because applying the classification algorithms to the uncompressed data resulted in significantly worse results. After normalization and PCA the data was ready for classification by the algorithms described in the following sections. The hyperparameters for each algorithm used in this work are presented in the supplementary material in the supplementary material

\subsection{ExtraTrees Classifier}

The ExtraTrees Classifier (ETC) is an extension of the popular Random Forests Classifier, which in turn is an ensemble of the Decision Tree (DT) algorithm~\cite{breiman_random_2001}. The main idea behind the DT algorithm is to construct a binary tree for the feature space of the data set. However, the algorithm  suffers from high training variance and overfitting. The ensemble of trees in the Random Forest Classifier is used to overcome these problems by generating a large, random set of such decision trees (hence named a random forest) and averages over their predictions to obtain the final prediction~\cite[p. 344]{James2021}. Trees are chosen such that they fit the underlying dataset optimally. The ETC, in contrast, aims at high variation amongst the decision trees even if it requires sacrificing accuracy of fit.
Random Forest-type algorithms are often treated as black-box methods~\cite{mashayekhi_rule_2015}.

In this work, the ETC algorithm from the scikit-learn library~\cite{scikit-learn} for Python was used with two parameters: \textit{n\_estimators} and \textit{criterion}. The parameter \textit{n\_estimators} specifies the number of trees in the ensemble. Choosing this parameter value is noncritical, and generally, a large number of trees is selected~\cite[p. 341]{James2021} with the default value being 100. The \textit{criterion} parameter accepts three values: \textit{gini}, \textit{entropy} and \textit{log\_loss}. For our application either \textit{gini} or \textit{entropy} would be good choices, since there is no considerable difference between them, except for computational complexity. \cite{raileanu_theoretical_2004} One of these three functions is needed for defining decision nodes in the tree. In this work, the function \textit{gini} was used for splitting the nodes. For the test, the optimal parameter set was selected using the grid search technique. The grid search technique is a method for finding the best combination of parameters for a classifier by simply iterating over all combinations of the predefined parameter grid. By means of the Out-Of-Bag score, it was found that 100 estimators were enough to achieve a cross-validated accuracy of 80\%. Other parameters were set to default values in the scikit-learn implementation.

\subsection{K Nearest Neighbors Classifier}

The widely known and simple K Nearest Neighbors (KNN) classifier is effective for many machine learning tasks. The idea behind this algorithm is to find K training points $x_{1,...,K}$ with known labels that are closest with respect to a distance measure to a query point $x_0$ without label. Based on the labels of the K neighbors a label for the query point is derived using majority vote~\cite[p. 463]{hastie_elements_2009}. Generally, this algorithm has two main parameters: the number of nearest neighbors K used to vote the predicted class and the distance metric. For the tests in this paper Euclidean distance was chosen. The problem of selecting the right number of closest neighbors is known as the model selection problem~\cite[p. 15]{kramer_dimensionality_nodate}. Selecting only one voting neighbor can result in overfitting \cite[p. 15]{kramer_dimensionality_nodate}. Eventually, three voting neighbors were selected based on the grid search algorithm.

Drawbacks of the KNN classifier include large memory requirement and its slowness for databases with a large number of training points if distances between query point and all labeled points are calculated, and its tendency to focus on irrelevant features. The latter weakness can be mitigated by applying a feature transformation or feature selection method to the dataset before running the KNN on it. For tackling the slowness of prestructuring, editing the data, or computing only partial distances are valid options \cite{hastie_elements_2009}. One prestructuring technique that has been successfully used in \cite{Muller2018} for classification of scents based on ion mobility spectrometry are k-dimensional trees \cite{bentley1975}, where k is not to be confused with the K from the KNN.

\subsection{Linear Discriminant Analysis}

Linear Discriminant Analysis (LDA) has demonstrated its effectiveness in many problems, including classification of DMS dispersion plots. For example, it has shown its ability to differentiate between two types of porcine tissue from surgical smoke measured by DMS with classification accuracy close to 93\% \cite{Kontunen2021}. The idea behind LDA is to find a decision boundary between classes of samples such that inter-class variability is maximized and intra-class variability is minimized. The advantages of LDA are its simplicity, that it provides a closed-form solution, and, its small number of hyperparameters. The two most important hyperparameter are \textit{solver} and \textit{shrinkage}. The solver parameter was set to \textit{lsqr}, which stands for Least Squares.
The \textit{shrinkage} parameter is a form of regularization that improves estimation of covariance matrices. In this work the shrinkage-parameter was set to \textit{auto}, which determines an optimal shrinkage parameter analytically according to the approach proposed in~\cite{ledoit_honey_2004}.

\subsection{Multilayer Perceptron}

A Multilayer Perceptron (MLP) is a non-linear machine learning algorithm that has the ability to approximate decision boundaries in case of complicated and highly non-linear problems~\cite[p. 225]{bishop_pattern_2006}. MLP is an extension of the ordinary binary perceptron algorithm. The idea behind the perceptron algorithm is to weigh inputs and pass the weighted inputs into a nonlinear function. The MLP consists of several layers of perceptrons, and is often called Artificial Neural Network. Usually, an MLP has at least three layers: input, hidden and output. The training is performed by backpropagating derivatives from the end of the network. For a more detailed explanation on MLP the reader is refered to~\cite{bishop_pattern_2006}.

\begin{figure*}
	\centering
	\begin{subfigure}{0.9\textwidth}
		\includegraphics[width=1\textwidth]{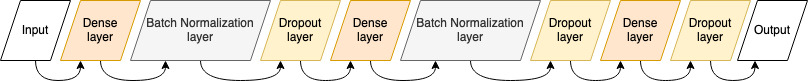}
		\caption{MLP}
		\label{fig:mlp_structure}
	\end{subfigure}
	\begin{subfigure}{0.9\textwidth}
	\includegraphics[width=1\textwidth]{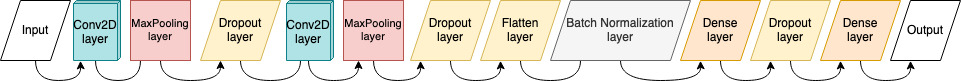}
	\caption{CNN}
	\label{fig:cnn_structure}
	\end{subfigure}
	\begin{subfigure}{0.9\textwidth}
		\includegraphics[width=1\textwidth]{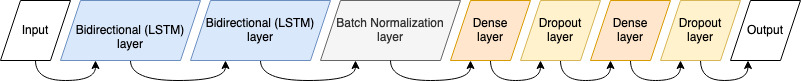}
		\caption{LSTM}
		\label{fig:lstm_structure}
	\end{subfigure}
\caption{Architectures of the artificial neural networks described in Section~\ref{sec:cmethods}}
\end{figure*}

The implemented architecture of MLP is shown in Figure \ref{fig:mlp_structure}. For finding the optimal parameter set grid search algorithms implemented specifically for neural networks were used. The network consisted of three hidden dense layers, each followed by a dropout layer and a batch normalization. The dense layers were initialized with the \textit{glorot\_uniform} initializer~\cite{glorot} and the layer regularization parameter was set to L2~\cite{Goodfellow-et-al-2016}. The dropout rate for all dropout layers was set to 0.1, which means that during the training 10\% of neurons were randomly disabled. The dropout approach enables the neural network to better generalize and avoid overfitting \cite{baldi_dropout_2014}.

\subsection{Convolutional Neural Network}

The Convolutional Neural Network (CNN) is a special kind of artificial neural network, which is intended for processing data that has a known grid-like topology. Such data can be images, which can be thought of as a 2-D grid of pixels~\cite{Goodfellow-et-al-2016}. CNNs have demonstrated their suitability for image classification, object recognition, etc. For example, Anttalainen et al. applied a CNN regression model for detecting the concentration of lecithin based on DMS measurements ~\cite{Anttalainen2021}.

The dispersion plots studied in this paper can be interpreted as images. Thus, it was hypothesized that employing CNN for the classification of dispersion plot may be successive. The used architecture of CNN is schematically shown in Figure \ref{fig:cnn_structure}. The employed CNN contained two convolution layers with 16 and 8 feature maps accordingly. Each convolutional layer was followed by MaxPooling and dropout layers. The end of the network contained two hidden dense layers.

Convolutional neural networks have many parameters to be tuned: number of convolution layers, kernels, activation functions, regularizers, architecture of the network, etc. Having enough computation capacity, it is very convenient to tune a neural network by iterating different combinations of the hyperparameters. For the tests in this paper, the optimal combination of hyperparameters was found by applying grid search.

\subsection{LSTM Neural Network}

Long Short Term Memory (LSTM) is another extension of artificial neural networks~\cite{hochreiter_long_1997}. Instead of neurons as in MLP or convolutions as in CNN, LSTM layers are used for capturing features, i.e. temporal patterns. The difference between a general neuron and LSTM neuron is that each LSTM neuron has a shortcut that retains information from previous time steps. Thus, the LSTM Neural Network can exploit historic information. The so-called bidirectional LSTM layers extend their functionality allowing reversing of the current time steps for capturing prior as well as posterior information.

A DMS dispersion plot represents a set of measurements performed with different settings. That is, the separation voltage increases from bottom to top and the compensation voltage increases from left to right. The different combinations of separation and compensation voltages are measured as sequences and can hence be interpreted as sequential information. Thus, it was hypothesized that LSTM neural networks are capable of capturing sequential features and yielding good classification accuracy. To our knowledge, there are no publications discussing applying sequential models to DMS dispersion plots, and we are providing the first results of experiments with LSTM.

Before entering the LSTM network, each dispersion plot was preprocessed by normalizing row-wise and clipping to the area containing the most entropy as described in Subsection~\ref{sec:prepro}. The clipping resulted in 20 samples with 100 features each.
The optimal set of parameters was found by applying grid search technique since there is no rule of thumb on how to define the number of neurons in the LSTM layer, learning rate, momentum, activation function, etc. The implemented architecture of the LSTM network is visualized in Figure \ref{fig:lstm_structure}. The LSTM neural network contained 8 bidirectional LSTM neurons on the first layer and 256 bidirectional neurons on the second LSTM layer. The two hidden dense layers contained 700 and 500 neurons, respectively. Additionally, the network had two dropout layers between the hidden layers with a dropout rate of 10\%. 

%% file: tex/results-philipp.tex
All algorithms were tested with 2 types cross-validation techniques. The first was Repeated Stratified 10-fold cross-validation. Stratified means that the algorithm maintains the same proportion of data in both training and testing splits. The repeated means that every cross-validation iteration is repeated N times and the average result is presented. The 10-fold means that the dataset is divided into ten subsets. The algorithm trains on nine splits and leave the tenth split for tests. This process is repeated for each subset. The second cross-validation technique was LeaveOneGroupOut cross-validation based on flowrates. This algorithm splits the dataset into subset based on groups. In this case the group was flow rate. On the first iteration the algorithm trains on flow rates 8, 16, 32, 64, and 128 was left for tests. On the next iteration the algorithm trains on 8, 16, 32, 128, and 64 was left for tests, etc.
Table \ref{tbl:results} shows cross-validation results for the Repeated Stratified 10-fold algorithm, and for LeaveOneGroupOut cross-validation.

\begin{figure*}[h]
	\centering
	\includegraphics[width=1\linewidth]{"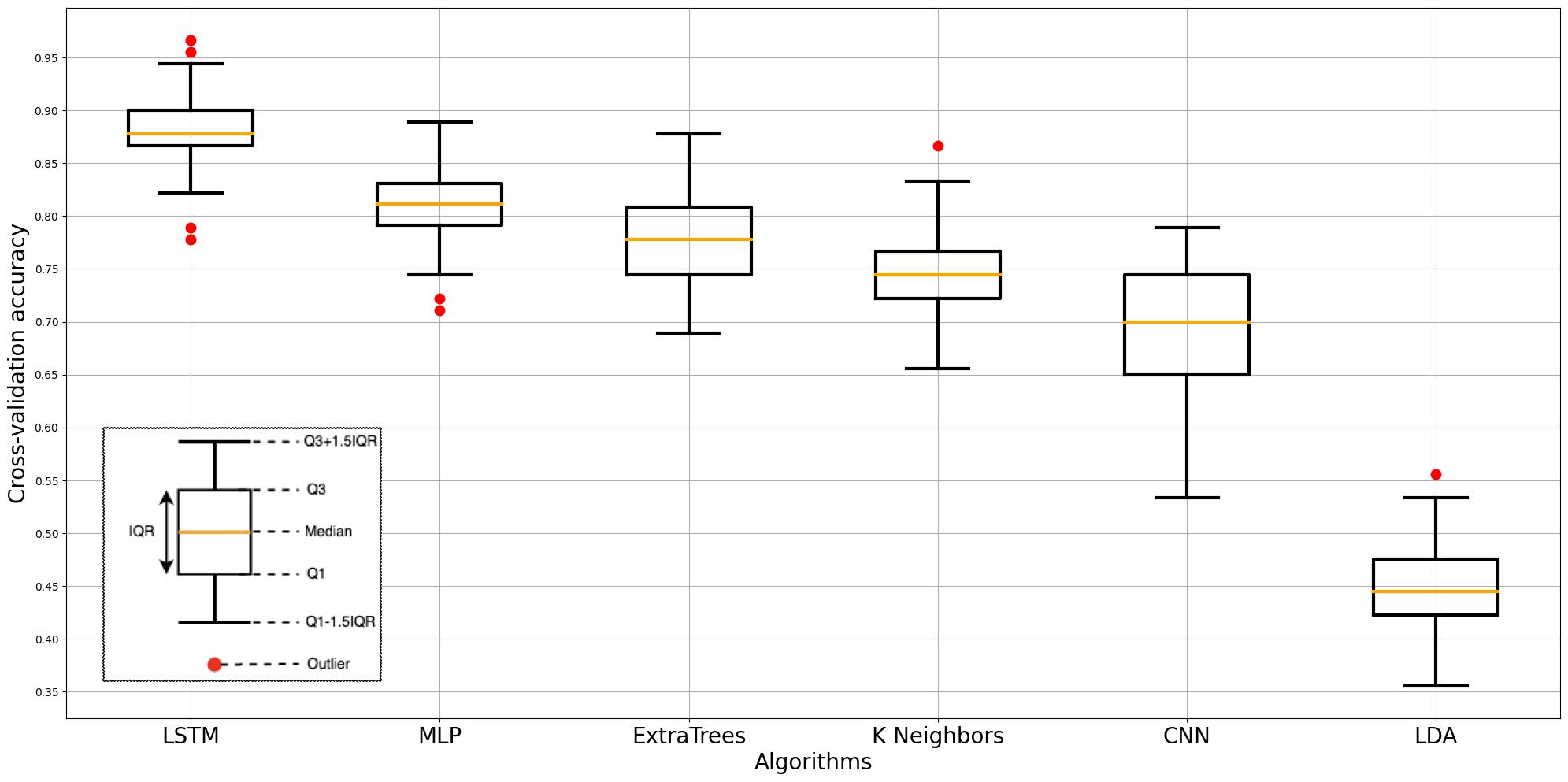"}
	\caption{Boxplots plot showing cross-validation results for each algorithm. The box extends from the first (Q1) to the third (Q3) quartile. The orange line depicts the median (the second quartile, Q2). The distance between Q1 and Q3 forms the interquartile range (IQR). All data points that lie beyond 1.5 times the IQR are considered to be outliers.}
	\label{fig:whiskersplot}
\end{figure*}

Classification accuracy was measured as the ratio of correctly classified chemicals.
Table~\ref{tbl:results} shows cross-validation results for $k$=10 and for the LeaveOneGroupOut cross-validation (GCV). Additionally, cross-validation results can be compared from the box plot (Figure \ref{fig:whiskersplot}). The boxes in the box plot span interquartile range (IQR = Q3 – Q1, 75th percentile – 25th percentile). The orange line in the box shows median (Q2, 50th percentile). The whiskers extend by 1.5IQR and the red dots beyond the whiskers are considered as outliers. The outlier data were not excluded from evaluation to not artificially increase the accuracy levels. Also, the accuracies from each cross-validation were collected and compiled into confusion matrices (see the supplementary material). Each entry in the confusion matrix contains average accuracy for a particular chemical and standard deviation over all cross-validation iterations.

During preliminary tests it was found that the hardest chemical to differentiate from the remaining four chemicals was methyl cyclopentenolone and the easiest was ethyl-2-methylbutyrate. As can be seen from Table \ref{tbl:results} the most misclassifications were encountered for 2PEtOH and MCP. This phenomenon is observed for all tested classification algorithms. Additionally, the E2MB have best classification scores compared to other chemicals. This can be explained by higher concentration and can be observed in Figure \ref{fig:e2mbexample}.

PCA was applied to the data prior to classification for all algorithms except CNN and LSTM. Since PCA transforms the original data into a subspace that changes the original meaning of the features, it is impossible to say what impact each of the original features had on the classification results for most of the algorithms. For more detailed information on performance of the classifiers the reader is referred to the supplementary material.

\input{tex/cv_table}

LDA was selected as baseline classifier due to its comparatively good classification accuracy in~\cite{Kontunen2021} and~\cite{haapalamethod2022}.The inferior performance of LDA compared to both ~\cite{haapalamethod2022} and~\cite{Kontunen2021} can be explained by the fact that in this paper non-binary classification was required, while in~\cite{haapalamethod2022} the classification problem was binary. Furthermore, resolutions and concentrations of the samples 
 differed considerably. Also, dimensionality in \cite{haapalamethod2022} was 1\,200 data point, and dispersion plots used in current work contained 8\,000 data points.

 The training and validation history of the neural networks is shown in the supplementary material. Each plot has both accuracy and loss curves for the training and validation sets.  The accuracy and loss curves demonstrate convergency of all three networks. Also, the behavior of the curves demonstrate that no overfitting occurred during training.


As can be seen, the LSTM neural network achieved the highest classification accuracy of 88.4\% and 91.0\% for 10 fold cross-validation and for the group cross-validation, which confirms that it is beneficial to interpret dispersion plots as series of measurements evolving sequentially. This allows the LSTM network to capture relations between measurements within different separation voltages and consequently yield considerably higher accuracies for Carvone, 2PEtOH and MCP than any other method. It furthermore achieves accuracies similar to the best methods for nBuOH and E2MB.

The LDA achieved the best accuracy in discriminating E2MB from other chemicals. This can be explained by the visual differences in E2MB dispersion plots (Figure~\ref{fig:dplotsexample}) compared to dispersion plots of the other chemicals.


%% file: tex/cv_table.tex
\begin{table*}[!ht]
	\centering
	\caption{The table contains classification results. Columns two,  and three show cross-validation accuracies with the standard deviations for cross-validation of fold size 10 (CV10) and LeaveOneGroupOut (GCV) in percents. Columns 4 to 8 contain average classification accuracies for each chemical with the standard deviations.}
	\label{tbl:results}
	\resizebox{\textwidth}{!}{%
		\begin{tabular}{c|cc|ccccc|}
			\cline{2-8}
			&
			\multicolumn{2}{c|}{\textbf{CV accuracy {[}\%{]}}} &
			\multicolumn{5}{c|}{\textbf{Accuracy for each chemical {[}\%{]}}} \\ \cline{2-8} 
			&
			\multicolumn{1}{c|}{\textbf{CV10}} & \textbf{GCV} &
			\multicolumn{1}{c|}{\textbf{nBuOH}} &
			\multicolumn{1}{c|}{\textbf{Carvone}} &
			\multicolumn{1}{c|}{\textbf{E2MB}} &
			\multicolumn{1}{c|}{\textbf{2PEtOH}} &
			\textbf{MCP} \\ \hline
			\multicolumn{1}{|c|}{\textbf{ETC}} &
			\multicolumn{1}{c|}{81.1 $\pm$3.6} & 82.5 $\pm{8.4}$ &
			\multicolumn{1}{c|}{87.7 $\pm$7.6} &
			\multicolumn{1}{c|}{79.0 $\pm$8.1} &
			\multicolumn{1}{c|}{95.6 $\pm$4.8} &
			\multicolumn{1}{c|}{74.9 $\pm$7.9} &
			71.8 $\pm$7.5 \\ \hline
			\multicolumn{1}{|c|}{\textbf{KNN}} &
			\multicolumn{1}{c|}{74.7 $\pm$3.9} & 75.8 $\pm{14.5}$ &
			\multicolumn{1}{c|}{81.1 $\pm$9.0} &
			\multicolumn{1}{c|}{74.3 $\pm$8.3} &
			\multicolumn{1}{c|}{93.9 $\pm$6.0} &
			\multicolumn{1}{c|}{68.8 $\pm$8.3} &
			60.8 $\pm$9.5 \\ \hline
			\multicolumn{1}{|c|}{\textbf{LDA}} &
			\multicolumn{1}{c|}{44.7 $\pm$4.9} & 46.3 $\pm{6.7}$ &
			\multicolumn{1}{c|}{50.2 $\pm$12.4} &
			\multicolumn{1}{c|}{33.1 $\pm$7.2} &
			\multicolumn{1}{c|}{\textbf{97.6 $\pm$4.9}} &
			\multicolumn{1}{c|}{41.6 $\pm$9.5} &
			21.5 $\pm$10.0 \\ \hline
			\multicolumn{1}{|c|}{\textbf{MLP}} &
			\multicolumn{1}{c|}{78.0 $\pm$4.0} & 79.6 $\pm{13.3}$ &
			\multicolumn{1}{c|}{84.9 $\pm$7.9} &
			\multicolumn{1}{c|}{75.6 $\pm$10.0} &
			\multicolumn{1}{c|}{96.8 $\pm$4.6} &
			\multicolumn{1}{c|}{75.1 $\pm$9.4} &
			65.5 $\pm$8.5 \\ \hline
			\multicolumn{1}{|c|}{\textbf{CNN}} &
			\multicolumn{1}{c|}{69.3 $\pm$6.1} & 75.4 $\pm{13.9}$ &
			\multicolumn{1}{c|}{74.5 $\pm$9.7} &
			\multicolumn{1}{c|}{67.8 $\pm$12.1} &
			\multicolumn{1}{c|}{89.9 $\pm$9.7} &
			\multicolumn{1}{c|}{64.3 $\pm$12.7} &
			57.1 $\pm$10.4 \\ \hline
			\multicolumn{1}{|c|}{\textbf{LSTM}} &
			\multicolumn{1}{c|}{\textbf{88.4 $\pm$3.5}} & \textbf{91.0 $\pm{9.5}$} &
			\multicolumn{1}{c|}{\textbf{91.3 $\pm$6.8}} &
			\multicolumn{1}{c|}{\textbf{86.7 $\pm$7.8}} &
			\multicolumn{1}{c|}{96.9 $\pm$4.3} &
			\multicolumn{1}{c|}{\textbf{85.6 $\pm$7.6}} &
			\textbf{84.9 $\pm$8.7} \\ \hline
		\end{tabular}%
	}
\end{table*}

%% file: tex/conclusions-philipp.tex
In this work, a new method for the classification of DMS dispersion plots was presented. The novel idea behind this method is to interpret dispersion plots as a set of measurements evolving sequentially. Hence, a time-series approach for classification was suggested. Following this idea, for the first time, to our knowledge, a LSTM model was applied to dispersion plots and achieved multilabel classification accuracy of 89\%. It was compared with other proven classification techniques, but none of them yielded the same or higher accuracy in either LeaveOneGroupOut or 10-fold cross-validation tests. Even when considering classification for single chemicals, LSTM consistently yielded the highest or close to highest accuracies.



In the future another approach for further improving the classification accuracy could be investigating the alpha-curves geometrically. The alpha-curves are the main descriptors of gas being measured. Thus, revealing alpha-curves and extracting their geometrical features can potentially increase the classification accuracy~\cite{rauhameri_alpha_curves}. However, DMS measurements are very sensitive to temperature and humidity changes due to their impact on the mobility of ionized molecules \cite[p.~250]{eiceman_ion_2013}. The state-of-the-art approach for dealing with the impact of environmental conditions is to normalize the measurement conditions, which is often insufficient. Therefore, in order to use the alpha-curves approach to its fullest potential, the impact of changes to environmental conditions on dispersion plots needs to be studied thoroughly.

In this paper the LSTM model was tested with dispersion plots of different chemicals. However, based on our experience with dispersion plots of other VOCs, such as sweat samples from Alzheimer and Parkinson patients, cancer tissues, etc., it is reasonable to assume that the LSTM method will work well with all types of dispersion plots.
